\documentclass[letterpaper, 10 pt, conference]{ieeeconf}  % Comment this line out if you need a4paper

\usepackage{graphicx}
\usepackage{cite}
\usepackage{amssymb}
\usepackage{amsmath} 
\usepackage[table,xcdraw]{xcolor} % 用于表格着色
\usepackage{multirow}
\usepackage{booktabs} % 提供更好的表格线条
\usepackage{hhline}
\usepackage{bbding}
\usepackage{url}
\IEEEoverridecommandlockouts                              % This command is only needed if 
\title{\LARGE \bf
TEM$^{3}$-Learning: Time-Efficient Multimodal Multi-Task Learning for Advanced Assistive Driving
}

\author{Wenzhuo Liu$^{1}$, Yicheng Qiao$^{2}$, Zhen Wang$^{1}$, Qiannan Guo$^{2}$, Zilong Chen$^{2}$, Meihua Zhou$^{2}$, \\ Xinran Li$^{3}$, Letian Wang$^{4}$, Zhiwei Li$^{5}$, Huaping Liu$^{2}$, Wenshuo Wang$^{1,*}$ % <-this % stops a space
\thanks{*denotes the corresponding author.}% <-this % stops a space
\thanks{$^{1}$Wenzhuo Liu, Zhen Wang, and Wenshuo Wang are with the Faculty of Marine Science and Technology, Beijing Institute of Technology, Zhuhai, China, 519088 (e-mail:wzliu@bit.edu.cn; 3220245402@bit.edu.cn; ws.wang@bit.edu.cn).}%
\thanks{$^{2}$Yicheng Qiao, Qiannan Guo, Zilong Chen, Meihua Zhou, and Huaping Liu are with the State Key Laboratory of Intelligent Technology and  Systems and Department of Computer Science and Technology, Tsinghua University, Beijing, 100084, China (e-mail: yichengqiao21@gmail.com; guoqiannan1203@163.com; chenzl22@mails.tsinghua.edu.cn; mhzhou0412@gmail.com; hpliu@tsinghua.edu.cn).}%
\thanks{$^{3}$Li Xinran is with the School of Engineering and Applied Science, Yale University, New Haven, CT, USA (e-mail: xinranll668@gmail.com).}%
\thanks{$^{4}$Letian Wang is with the University of Toronto, Toronto, M5S2E8, Canada (e-mail: lt.wang@mail.utoronto.ca).}%
\thanks{$^{5}$Zhiwei Li is with Beijing University of Chemical Technology, Beijing, 100029, China (e-mail: 2022500066@buct.edu.cn).}%
}

\begin{document}

\maketitle
\thispagestyle{empty}
\pagestyle{empty}

%%%%%%%%%%%%%%%%%%%%%%%%%%%%%%%%%%%%%%%%%%%%%%%%%%%%%%%%%%%%%%%%%%%%%%%%%%%%%%%%
\begin{abstract}

Multi-task learning (MTL) can advance assistive driving by exploring inter-task correlations through shared representations. However, existing methods face two critical limitations: single-modality constraints limiting comprehensive scene understanding and inefficient architectures impeding real-time deployment. This paper proposes TEM$^{3}$-Learning (\underline{T}ime-\underline{E}fficient \underline{M}ulti\underline{m}odal \underline{M}ulti-task Learning), a novel framework that jointly optimizes driver emotion recognition, driver behavior recognition, traffic context recognition, and vehicle behavior recognition through a two-stage architecture. The first component, the mamba-based multi-view temporal-spatial feature extraction subnetwork (MTS-Mamba), introduces a forward-backward temporal scanning mechanism and global-local spatial attention to efficiently extract low-cost temporal-spatial features from multi-view sequential images. The second component, the MTL-based gated multimodal feature integrator (MGMI), employs task-specific multi-gating modules to adaptively highlight the most relevant modality features for each task, effectively alleviating the negative transfer problem in MTL. Evaluation on the AIDE dataset, our proposed model achieves state-of-the-art accuracy across all four tasks, maintaining a lightweight architecture with fewer than 6 million parameters and delivering an impressive 142.32 FPS inference speed. Rigorous ablation studies further validate the effectiveness of the proposed framework and the independent contributions of each module. The code is available on \url{https://github.com/Wenzhuo-Liu/TEM3-Learning}.

\end{abstract}

%%%%%%%%%%%%%%%%%%%%%%%%%%%%%%%%%%%%%%%%%%%%%%%%%%%%%%%%%%%%%%%%%%%%%%%%%%%%%%%%
\section{INTRODUCTION}

Advanced Driver Assistance Systems (ADAS) can improve driving safety by continuously monitoring the driver’s state and traffic environment \cite{gong2023sifdrivenet,zhang2023oblique}. However, existing research is mostly limited to the solution of single tasks, such as driver emotion/behavior, traffic environment recognition, without addressing the inherent interdependencies between these tasks \cite{qian2019dlt}. Studies show a significant coupling relationship between the drivers’ state and the traffic environment \cite{yang2023aide,martin2019drive}. For example, drivers frequently adjust their behavior, such as changing lanes based on surrounding traffic conditions, while traffic congestion can induce driver anxiety. Integrating driver state and traffic environment recognition into a unified multi-task learning (MTL) framework can provide a more holistic understanding of driving scenarios and enhance the safety performance of ADAS \cite{yang2023aide}.

\begin{figure}
\centering
\includegraphics[width=0.49\textwidth]{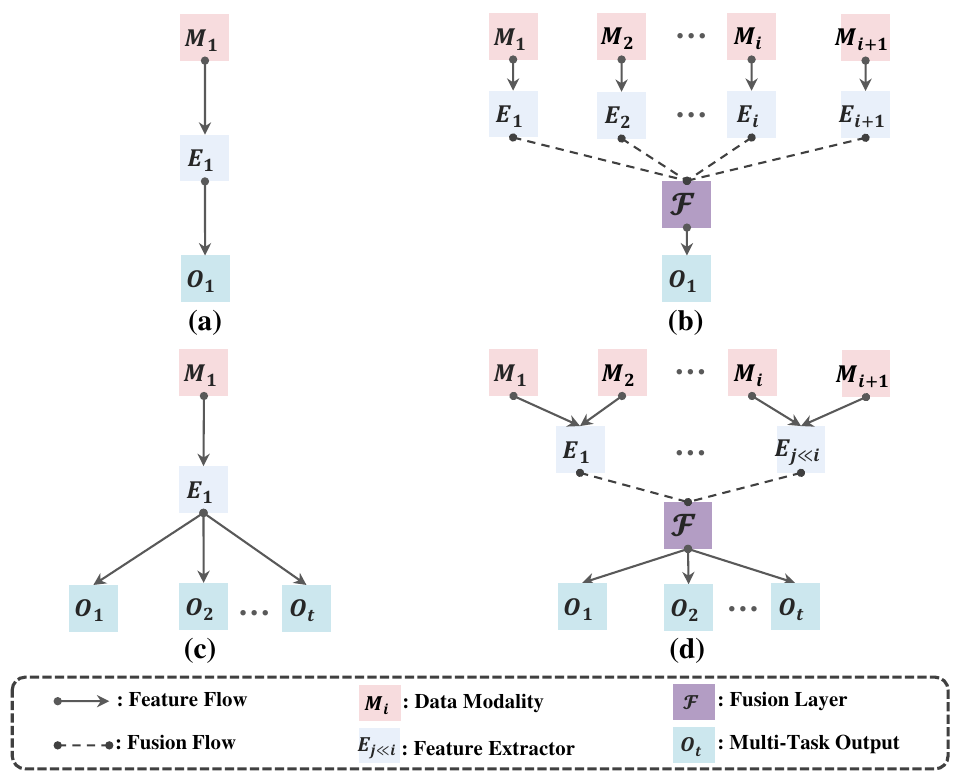}
\vspace{-1em}
\caption{Comparison of four mainstream algorithm frameworks: (a) single-modal single-task, (b) multimodal single-task, (c) single-modal multi-task, and (d) the proposed multimodal multi-task. As the number of tasks and modalities increases, model complexity rises significantly, making it crucial to balance computational efficiency with accuracy in multimodal multi-task learning.}
\label{fig1}
\vspace{-1em}
\end{figure}

Compared to single-task learning, MTL reduces overfitting by sharing features across tasks, thereby improving the performance of each individual task \cite{chowdhuri2019multinet,ishihara2021multi}. In the context of ADAS, existing MTL models generally focus on related sub-tasks. For example, Wu proposed a model that jointly learns multiple traffic environment-related tasks, enabling simultaneous recognition of lane markings, drivable areas, and object detection in driving scenes \cite{wu2022yolop}. Similarly, Xing et al. focus on driver-related MTL to recognize the driver’s emotions and behaviors \cite{xing2021multi}. While these approaches have yielded performance improvements, they often struggle with negative transfer, where task performance deteriorates due to task-related conflicts or differences\cite{liu2019loss}.

MTL models tend to have more complex structures compared to single-task learning, as they must simultaneously handle multiple tasks \cite{wu2022yolop,zhan2024yolopx} (Fig. \ref{fig1} (a) and (c)). Many MTL methods rely on inputs from a single modality such as driving scene images \cite{choi2024multi,chen2023umt}. However, optimal task performance requires complementary multimodal inputs\cite{guo2023temporal,liu2024glmdrivenet}. For example, combining driver images with eye movement data can improve emotion recognition \cite{mou2023driver}, while fusing image and point cloud data can enhance traffic object recognition \cite{hao2024coarse,alaba2024emerging}. Unfortunately, many multimodal input algorithms still rely on independent feature extraction \cite{wang2023openoccupancy,liu2025umd,liu2024fmdnet,huang2024mfe}, which significantly increases model parameters, reduces inference speed, and compromises practical scalability (Fig. \ref{fig1} (b)). Therefore, a key challenge lies in effectively leveraging multimodal information while optimizing individual task performance, enhancing task-level synergy, and maintaining high inference speed with a minimal parameter count.

To address these challenges, we propose a Time-efficient multimodal multi-task learning network that simultaneously performs four tasks --- driver emotion recognition (\textsc{der}), driver behavior recognition (\textsc{dbr}), traffic context recognition (\textsc{tcr}), and vehicle behavior recognition (\textsc{vbr}) --- using multimodal data (Fig. \ref{fig1} (d)). This is achieved through two key components: the Mamba-based multi-view temporal-spatial feature extraction subnetwork (MTS-Mamba) and the multi-task learning-based gated multimodal feature integrator (MGMI). MTS-Mamba introduces a forward-backward temporal scanning mechanism and a global-local spatial feature extraction strategy, enabling efficient extraction of temporal-spatial features from multi-view sequential images at low computational cost. This approach provides richer and more robust feature representations for subsequent multi-task recognition. MGMI, inspired by previous work \cite{ma2018modeling}, incorporates a multi-gating mechanism that dynamically adjusts the weights of modality features based on task-specific attention. This selective reinforcement of important features helps alleviate negative transfer. We validated the effectiveness of our method using the publicly available AIDE dataset. Experimental results show that our model outperforms previous methods on all four tasks, with fewer than 6M parameters and exceptionally fast inference speed, demonstrating its efficiency and practicality. Our contributions include:

\begin{itemize}
    \item [$\bullet$] A time-efficient multimodal MTL framework that provides a new paradigm for multimodal MTL in ADAS.
    \item [$\bullet$] The MTS-Mamba subnetwork, which effectively extracts temporal-spatial features from multi-view sequential images across multiple dimensions.
    \item [$\bullet$] The MGMI mechanism, which adaptively adjusts attention to different modalities for each task, mitigating negative transfer and enhancing task-specific feature extraction.
\end{itemize}

\section{Related Work}

\subsection{Multimodal Learning}

ADAS systems that rely solely on single-modality data often struggle to handle the complex and dynamic challenges encountered during driving \cite{mou2023driver,gong2022multi,gan2024segmentation,shi2023bssnet,wang2023path,bi2025vm,alaba2024emerging}. Each modality offers unique advantages and limitations, and leveraging multiple modalities can provide a more comprehensive understanding of the driving environment.  For example, combining multi-view driving scene images with LiDAR data enhances environmental perception \cite{xu2021fusionpainting,li2024mipd}. Similarly, integrating driver images and joint information can effectively capture facial expressions and behavioral states \cite{yang2023aide}. As a result, many studies have increasingly turned to multimodal data to enhance task accuracy. For instance, Zhou et al. \cite{zhou2020driving} used front-view driving scene images, driver images, and vehicle speed data for driver behavior recognition, while Liu et al. \cite{liu2022epnet++} combined camera images and LiDAR data for 3D object detection.

However, multimodal learning in ADAS faces two key challenges. First, many existing models rely on independent feature extraction branches for each modality, even when processing multi-view images \cite{liu2024fmdnet,guo2023temporal,tan2025graph,zhou2020driving,liu2025mmtl}. This approach significantly increases the model’s parameter count and overlooks potential intermodal interactions, leading to inefficient information use. Second, multimodal fusion methods are often limited to combining only a few modality features, restricting their generalization ability and scalability.

\subsection{Multi-task Learning}

MTL improves model performance on each task by sharing features across tasks \cite{chowdhuri2019multinet,ishihara2021multi}. The two predominant strategies for parameter sharing are hard and soft parameter sharing. In hard parameter sharing, most parameters are shared across tasks, with task-specific differentiation occurring only in the final layers \cite{li2020knowledge,cui2024textnerf,cao2023relational}. For example, Wu et al. \cite{wu2022yolop} adopted this structure to simultaneously achieve traffic object detection, drivable area segmentation, and lane detection. While simple and efficient, this approach is prone to negative transfer when tasks exhibit substantial differences, limiting its generalizability. To mitigate this, soft parameter sharing has been proposed, where each task retains independent parameters but leverages shared features, thus preserving task independence and reducing task conflicts \cite{gao2023enhanced,chen2023adamv,yang2022cross}. For instance, Choi et al. \cite{choi2024multi} designed a task-adaptive attention generator to enable independent parameters for tasks like monocular 3D object detection, semantic segmentation, and dense depth estimation.

Although soft parameter sharing alleviates task conflicts, two critical challenges remain. First, introducing task-specific parameters increases the model's flexibility but significantly raises the parameter count, which can hinder real-time performance --- a major concern in applications like ADAS, where real-time processing is crucial. Second, existing MTL models in ADAS are predominantly designed for single-modality inputs (e.g., driving scene images), limiting their applicability to multimodal scenarios and constraining the full potential of MTL-based solutions.

\begin{figure}[h]
    \centering
    \includegraphics[width=0.49\textwidth]{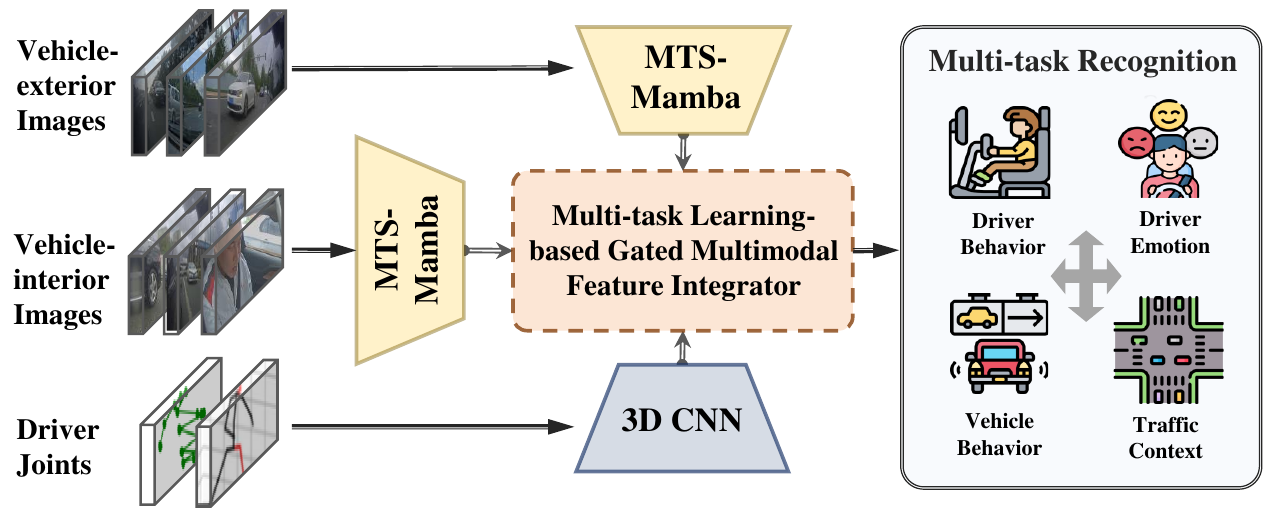}
    \vspace{-1em}
    \caption{The overall pipeline of TEM$^{3}$-Learning. MTS-Mamba and 3D CNN are used to extract multimodal features from vehicle-exterior images, vehicle-interior images, and driver joints, respectively. The multi-task learning-based gated multimodal feature integrator (MGMI) adaptively fuses these features, enabling multi-task recognition.}
    \label{fig2}
\end{figure}

\section{Methodology}
This section presents the overall structure and key modules of the proposed TEM$^3$-Learning network. We first describe the overall architecture of the network and then detail two core modules: MTS-Mamba and MGMI.

\subsection{Network Overview}
\label{Network Overview}
To fully leverage the synergies between multimodal and multi-view data while balancing accuracy and efficiency in a MTL context, our proposed TEM$^{3}$-Learning network (Fig. \ref{fig2}) consists of three core components: the multi-branch feature extraction layer, MGMI, and the multi-task recognition layer.

The first component, the multi-branch feature extraction layer, processes multimodal data through two modules: MTS-Mamba and 3D CNN. MTS-Mamba extracts deep temporal-spatial features from vehicle-exterior images (front, left, and right views) and vehicle-interior images (inside-view, driver’s facial and body images). The 3D CNN module focuses on extracting high-level semantic features from the driver’s posture and gestures. These extracted features are then integrated by MGMI, which first uses a self-attention mechanism to obtain task-shared features, followed by a multi-gating mechanism to adaptively weight features based on task-specific attention, thus improving feature extraction and alleviating conflicts between tasks.

During training, we use a cross-entropy loss function that integrates individual task losses to optimize the overall model performance. The total loss $L_{\text{total}}$ is computed as:

\begin{equation}
L_{\text{total}} = \sum_{r=1}^{m} \text{CrossEntropy}\Big( \hat{y}_{r}, y_{r} \Big),
\end{equation}
where $\hat{y}_{r}$ denotes the recognition results for task $r$, and $y_{r}$ denotes the corresponding ground truth. The number of tasks $m$ is set to 4, corresponding to the four tasks: driver emotion recognition, driver behavior recognition, traffic context recognition, and vehicle behavior recognition.

\subsection{Mamba-based Multi-view Temporal-spatial Feature Extraction Subnetwork (MTS-Mamba)}
\label{MTS-Mamba}
Multi-view sequential images from both vehicle interiors and exteriors exhibit strong temporal-spatial correlations. For instance, sequential vehicle-interior images effectively capture driver behavior (e.g., looking around, making calls), while multi-view vehicle-exterior images provide a comprehensive understanding of surrounding environments (e.g., pedestrians, vehicles, obstacles). Leveraging these temporal-spatial features is crucial for enhancing ADAS performance in environmental perception and behavior recognition. However, in real-world driving scenarios --- especially in complex scenarios like congested intersections where dynamic targets and complex movement patterns are prevalent --- modeling these features becomes increasingly challenging. The difficulty arises from the drastic environmental changes and the need to balance modeling quality with real-time performance. CNNs offer good real-time performance but are limited by their local receptive field, which hampers the capture of global temporal-spatial features and, consequently, task recognition accuracy. On the other hand, attention-based methods, while capable of global modeling, suffer from high computational complexity, making them unsuitable for processing large volumes of multi-view sequential data in real-time ADAS applications. To address these challenges, we propose MTS-Mamba, which combines a dual-path temporal-spatial feature extraction structure with a State Space Model (SSM) based on Mamba \cite{gu2023mamba}. This approach efficiently captures multi-view sequential image features while maintaining low computational cost, ensuring a balance between recognition accuracy and real-time performance.

\begin{figure*}[h]
    \centering
    \includegraphics[width=0.98\textwidth]{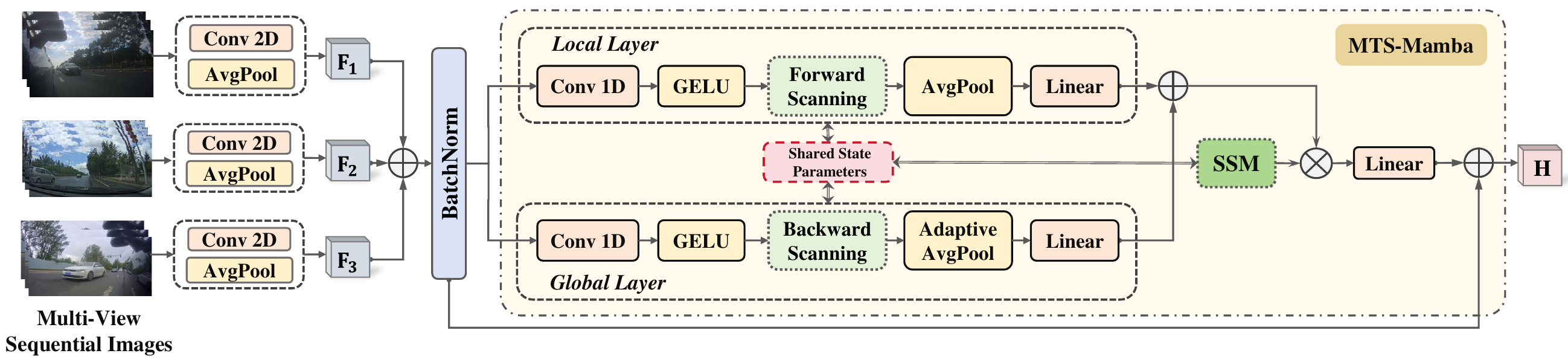}
    \caption{The structure diagram of MTS-Mamba, which includes the forward and backward scanning mechanisms, as well as the global-local spatial feature extraction strategy.}
    \label{fig:GCFANet}
    \vspace{-1em}
\end{figure*}

Before MTS-Mamba, we need to process multi-view sequential images. First, the sequential images of the j-th view are concatenated along the channel dimension, forming a sequence $\mathbf{I}_j$. Each $\mathbf{I}_j$ is then processed through deep convolution and adaptive average pooling to extract initial features of identical dimensions. These features are integrated into a feature map $\mathbf{F}_f \in \mathbb{R}^{C \times H \times W}$ , where $C$, $H$, and $W$ represent the channel count, height, and width, respectively. This feature map is subsequently input into the MTS-Mamba module for further processing.

For the input feature map $\mathbf{F}_f$, we first apply 1D convolution with GELU activation to enhance feature representation within the dual-path temporal-spatial feature extraction structure,i.e.,  global and local layers. The local and global layers employ forward and backward scanning, respectively, to extract bidirectional temporal features. Specifically, given that $\mathbf{F}_f$ spans 16 consecutive frames, the local layer perform a forward scan where each channel of $\mathbf{F}_f$ is linearly transformed by the state parameters $\mathbf{B} \cdot \mathbf{C}^\top$ of the state-space model, capturing forward temporal dependencies. In the global layer, the spatial order of $\mathbf{F}_f$ is reversed during the backward scanning to capture temporal interdependencies from both directions. Then, it is linearly transformed with $\mathbf{B} \cdot \mathbf{C}^\top$. During training, the shared state parameters $\mathbf{B}$ and $\mathbf{C}$ are updated via backpropagation to retain bidirectional temporal information. The SSM calculates the temporal feature weight $\mathcal{W}_{\text{ssm}}$ as:

\begin{equation}
\left\{
\begin{aligned}
\mathbf{B}^{(s+1)} = \mathbf{B}^{(s)} - \eta \cdot \frac{\partial \mathcal{L}}{\partial \mathbf{B}^{(s)}},
\\
\mathbf{C}^{(s+1)} = \mathbf{C}^{(s)} - \eta \cdot \frac{\partial \mathcal{L}}{\partial \mathbf{C}^{(s)}},
\end{aligned}
\right.
\end{equation}

\begin{equation}
\mathcal{W}_{\text{ssm}} = \sigma \left( \mathbf{A}\cdot \mathbf{d}_{\text{state}} + \mathbf{B} \cdot \mathbf{C}^{\top} \cdot \mathbf{d}_{\text{dim}} + \mathbf{D} \right),
\end{equation} 
where $\mathbf{B}^{(s+1)}$ denotes the updated shared weight parameter, $\eta$ is the learning rate, $\sigma(\cdot)$ is the sigmoid function, $\mathbf{A} \in \mathbb{R}^{C \times n}$ is the state transition matrix where $n$ indicates the state dimension, $\mathbf{d}_{\text{state}}$ and $\mathbf{d}_{\text{dim}}$ are the state unit vector and channel unit vector, and $\mathbf{D} \in \mathbb{R}^{C}$ is the bias vector. 

After capturing the bidirectional temporal features, the local layer uses 3$\times$3 average pooling and linear projection to extract local spatial features $\mathbf{F}_l$, while the global layer uses adaptive average pooling and linear projection to capture global spatial features $\mathbf{F}_g$. This captures detailed features and global context from spatial information at different scales, providing more comprehensive and rich spatial feature representations. We then merge the multi-scale spatial features $\mathbf{F}_l$ and $\mathbf{F}_g$ and multiply them by the temporal feature weight information $\mathcal{W}_{\text{ssm}}$, combining temporal dynamics with spatial structure. A residual connection is then used to combine the merged features with the original input features $\mathbf{F}_f$, producing the final output feature $\mathbf{F}_o$:

\begin{equation}
\mathbf{F}_{o} = \mathbf{F}_{f} + \gamma\cdot \text{LN}\big(\mathcal{W}_{\text{ssm}} \odot (\mathbf{F}_{l} + \mathbf{F}_{g}) \big),
\end{equation}
where $\gamma$ is the scaling factor, $\text{LN}$ denotes linear layer, and $\odot$ denotes element-wise multiplication. 

In summary, MTS-Mamba's unique dual-path temporal-spatial feature extraction structure effectively captures both multi-scale spatial features and bidirectional temporal dependencies. By introducing forward-backward scanning mechanisms, MTS-Mamba achieves a balance between accuracy in temporal-spatial feature modeling and computational efficiency, significantly enhancing real-time performance in ADAS applications.

\subsection{Multi-task Gated Multimodal Integrator (MGMI)}

We propose the Multi-task Gated Multimodal Integrator (MGMI), which introduces task-specific gating mechanisms to dynamically adjust the fusion weights of different modality features, enabling task-driven feature selection and alleviating task conflicts. Specifically, we first concatenate the features $\mathbf{H}_1$, $\mathbf{H}_2 \in \mathbb{R}^{C \times H \times W}$ extracted by MTS-Mamba with the features $\mathbf{H}_3$ from the 3D CNN module along the channel dimension to obtain the initial fused feature. This fused feature undergoes three separate convolution operations to produce query ($\mathbf{Q}$), key ($\mathbf{K}$), and value ($\mathbf{V}$), respectively. The computation process is as follows:

\begin{equation}
\mathbf{Q},\mathbf{K},\mathbf{V} = \mathcal{W}_{q,k,v}\big(\text{Concat}(\mathbf{H}_1,\mathbf{H}_2,\mathbf{H}_3)\big),
\end{equation}
The attention matrix is computed by performing a dot product between $\mathbf{Q}$ and $\mathbf{K}$, and multiplying the resulting attention matrix by $\mathbf{V}$ to obtain the weighted attention scores, which are reshaped back to $\mathbb{R}^{C \times H \times W}$ to yield the task-shared features.

To adapt to each task’s specific focus, we design a multi-gating mechanism. The task-shared features are input into four task-specific gating units, one for each task, which perform convolution, BatchNorm, and Sigmoid operations to calculate attention weights for the multimodal features $\mathbf{H}_1$, $\mathbf{H}_2$, and $\mathbf{H}_3$. A weighted sum of these features is calculated to produce the task-specific feature $\mathbf{F}_r$, where $r \in \{1,2,3,4\}$ represents each task. The task-specific feature $\mathbf{F}_r$ is

\begin{equation}
\mathbf{F}_{r} = \sum_{i=1}^{3} \mathbf{H}_{i} \odot \sigma_{r}^{i}\left(\text{BN}\big(\text{Conv2D}\big(\text{softmax}(\frac{\mathbf{Q} \cdot \mathbf{K}^\top}{\sqrt{d}}) \cdot \mathbf{V}\big)\big) \right),
\end{equation}
where $\sigma_{r}^{i}$ is the $i$-th gating Sigmoid function for task $r$, and $\text{BN}$ denotes the batch normalization operation. This design optimizes the feature fusion process for each task by dynamically adjusting the importance of each modality’s features, allowing the model to extract task-specific features while retaining shared characteristics. By alleviating negative transfer across tasks, MGMI enhances the performance of each task within the multi-task learning framework. Once multimodal features are fused for each task, the multi-task recognition layer applies independent pooling for each task’s feature $\mathbf{F}_r$ and inputs them into their respective classifiers, producing the prediction $\hat{y}_{r}$ for each task.

\label{MGMI}

\begin{figure}
\centering
\includegraphics[width=0.48\textwidth]{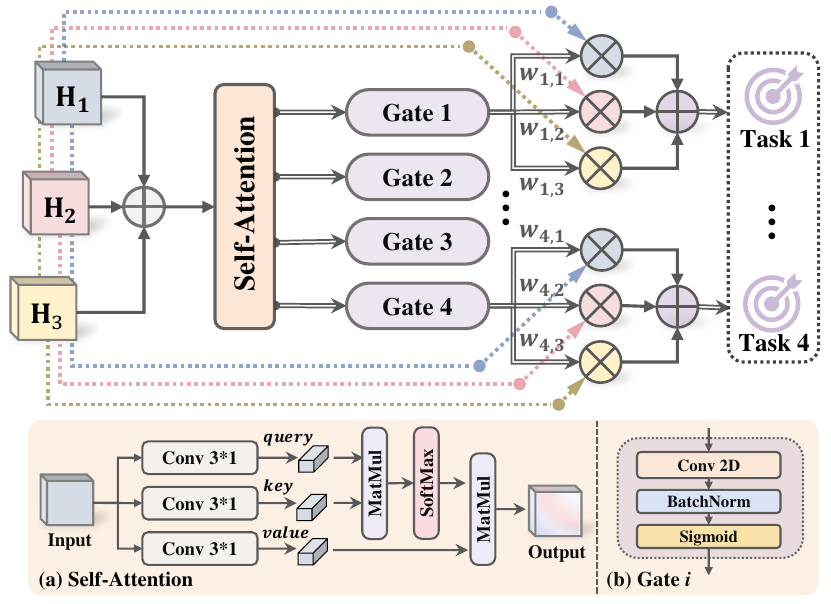}
\vspace{-1em}
\caption{Structure of the multi-task learning-based gated multimodal feature integrator (MGMI).}
\vspace{-1em}
\label{SME}
\end{figure}

\section{Experiments}

\subsection{Dataset}
The AIDE dataset is an open-source collection designed to advance ADAS research, consisting of 2,898 samples of time-series multimodal data, including multi-view images and driver joint positions. The multi-view images are captured from four perspectives: front-view, left-view, right-view, and inside-view. Each sample is annotated with labels for four tasks: driver emotion recognition (\textsc{der}), driver behavior recognition (\textsc{dbr}), traffic context recognition (\textsc{tcr}), and vehicle behavior recognition (\textsc{vbr}). The dataset is split into training, testing, and validation sets with proportions of 65\%, 15\%, and 20\%, respectively, to ensure robust evaluation.

\subsection{Data Preprocessing} 
The preprocessing pipeline follows three sequential steps: extraction of driver facial and body images, data synchronization, and data augmentation. First, driver facial and body regions are cropped using bounding box coordinates from inside-view images. The multimodal data (images and joint positions) is synchronized at 16 frames per second, ensuring temporal alignment of features across modalities. Each input sequence is formed from 16 consecutive frames to capture temporal dependencies. To further augment the dataset, multi-view images are subjected to random horizontal and vertical flips (50\% probability), enhancing the diversity of the training set and improving model robustness.

\begin{table*}[t]
\setlength{\tabcolsep}{4pt}
\centering
\renewcommand{\arraystretch}{1.2}
\vspace{-1em}
\caption{Comparison results with state-of-the-art algorithms on the AIDE dataset. The best results are marked in \textbf{bold}, and the second-best results are \cellcolor{blue!15} \raisebox{0.8ex}{\fcolorbox{blue!15}{blue!15}{\hspace{0.4cm}}}, and our method is highlighted with \cellcolor{gray!15} \raisebox{0.8ex}{\fcolorbox{gray!15}{gray!15}{\hspace{0.4cm}}}. P(M) represents the number of parameters in Millions, TE represents temporal embedding, TransE represents Transformer Encoder~\cite{vaswani2017attention}.}
\resizebox{\linewidth}{!}{%
\begin{tabular}{c|ccc|c|c|c|c|c|c|c}
\toprule
 & \multicolumn{3}{c|}{Backbone} & \multicolumn{4}{c|}{$\alpha_{\text{acc}} (\%) \uparrow $} & \multirow{2}{*}{\centering \begin{tabular}[c]{@{}c@{}}$\beta_{\text{macc}}$ \\ $(\%) \uparrow$\end{tabular}} & \multirow{2}{*}{\centering FPS $\uparrow$} & \multirow{2}{*}{\centering P(M) $\downarrow$} \\ \cline{2-8}
\multirow{-2}{*}{\centering Pattern} & \raisebox{-0.6ex}{\centering Multi-view Scene Images} & \raisebox{-0.6ex}{\centering Driver Images} & \raisebox{-0.6ex}{\centering Joints} & \raisebox{-0.6ex}{\centering \textsc{der}} & \raisebox{-0.6ex}{\centering \textsc{dbr}} & \raisebox{-0.6ex}{\centering \textsc{tcr}} & \raisebox{-0.6ex}{\centering \textsc{vbr}} & & \\ 
\midrule

\multirow{4}{*}{2D} & VGG16~\cite{simonyan2014very} & VGG16~\cite{simonyan2014very} & 3DCNN & 69.12 & 64.57 & 84.77 & 74.08 & 73.15 & 52.70 & 127.48 \\
 & Res18~\cite{he2016deep} & Res18~\cite{he2016deep} & 3DCNN & 68.78 & 64.33 & 89.76 & 78.59 & 75.37 & 57.30 & 107.77 \\
 & CMT~\cite{Guo_2022_CVPR} & CMT~\cite{Guo_2022_CVPR} & 3DCNN & 68.75 & 68.75 & \cellcolor{blue!15}93.75 & \cellcolor{blue!15}81.38 & 78.16 & 61.72 & 72.33 \\
 & GLMDriveNet~\cite{liu2024glmdrivenet} & GLMDriveNet~\cite{liu2024glmdrivenet} & 3DCNN & 71.38 & 66.57 & 90.23 & 77.19 & 76.34 & \cellcolor{blue!15}85.17 & 78.17 \\ \midrule
 \multirow{5}{*}{\begin{tabular}[c]{@{}c@{}}2D +\\ Timing\end{tabular}}& PP-Res18+TransE~\cite{vaswani2017attention} & Res18/34~\cite{he2016deep}+TransE~\cite{vaswani2017attention} & MLP+TE & 70.83 & 67.32 & 90.54 & 79.97 & 77.17 & -&- \\
 & Res34~\cite{he2016deep}+TransE~\cite{vaswani2017attention} & Res18/34~\cite{he2016deep}+TransE~\cite{vaswani2017attention} &MLP+TE & 72.65 & 67.08 & 86.63 & 78.46 & 76.21 &- &- \\
 & Res50~\cite{he2016deep}+TransE~\cite{vaswani2017attention} &Res34/50~\cite{he2016deep}+TransE~\cite{vaswani2017attention} &MLP+TE & 70.24 & 65.65 & 82.57 & 77.29 & 73.94 &- &- \\
 & VGG16~\cite{simonyan2014very}+TransE~\cite{vaswani2017attention} &VGG13/16~\cite{simonyan2014very}+TransE~\cite{vaswani2017attention} & MLP+TE & 71.12 & 67.15 & 85.13 & 78.58 & 75.50 &- &- \\
 & VGG19~\cite{simonyan2014very}+TransE~\cite{vaswani2017attention} & VGG16/19~\cite{simonyan2014very}+TransE~\cite{vaswani2017attention} & MLP+TE & 69.46 & 65.48 & 85.74 & 77.91 & 74.65 & -&- \\ \midrule
 \multirow{10}{*}{3D} & 3D-Res34~\cite{hara2018can} & 3D-Res34~\cite{hara2018can} & 3DCNN & 69.13 & 63.05 & 87.82 & 79.31 & 74.83 & 12.67 & 303.10\\
 & MobileNet-V1-3D~\cite{howard2017mobilenets} & MobileNet-V1-3D~\cite{howard2017mobilenets} & ST-GCN & 72.23 & 64.20 & 88.34 & 77.83 & 75.65 & 33.15 & 54.05 \\
 & MobileNet-V2-3D~\cite{sandler2018mobilenetv2} & MobileNet-V2-3D~\cite{sandler2018mobilenetv2} & ST-GCN & 68.47 & 61.74 & 86.54 & 78.66 & 73.85 & 12.16 & 83.78 \\
 & ShuffleNet-V1-3D~\cite{zhang2018shufflenet} & ShuffleNet-V1-3D~\cite{zhang2018shufflenet} & ST-GCN & 72.41 & \cellcolor{blue!15}68.97 & 90.64 &80.79 & \cellcolor{blue!15}78.20 & 51.29 & \cellcolor{blue!15} 31.49 \\
 & ShuffleNet-V2-3D~\cite{ma2018shufflenet} & ShuffleNet-V2-3D~\cite{ma2018shufflenet} & ST-GCN & 70.94 & 64.04 & 89.33 & 78.98 & 75.82 & 50.53 & 35.09 \\
 & C3D~\cite{tran2015learning} & C3D~\cite{tran2015learning} & ST-GCN & 63.05 & 63.95 & 85.41 & 77.01 & 72.36 & 25.62 & 158.46 \\
 & I3D~\cite{carreira2017quo} & I3D~\cite{carreira2017quo} & ST-GCN & 70.94 &66.17 & 87.68 & 79.81 & 76.15 & - & - \\
 & SlowFast~\cite{feichtenhofer2019slowfast} & SlowFast~\cite{feichtenhofer2019slowfast} & ST-GCN & 72.38 & 61.58 & 86.86 & 78.33 & 74.79 & -&-\\
 & TimeSFormer~\cite{bertasius2021space} & TimeSFormer~\cite{bertasius2021space} & ST-GCN & \cellcolor{blue!15} 74.87 & 65.18 & 92.12 & 78.81 & 77.75 & 25.63 & 158.46\\
 & Video Swin Transformer~\cite{liu2022video} & Video Swin Transformer~\cite{liu2022video} & 3DCNN & 73.44 & 65.63 & \cellcolor{blue!15}93.75 & 75.00 & 76.96 & 11.45 & 119.80\\ \midrule
 \rowcolor{gray!15}
\multirow{1}{*}{\textbf{Ours}} & \textbf{MTS-Mamba} & \textbf{MTS-Mamba} & \textbf{3DCNN} & \textbf{75.00} & \textbf{69.31} & \textbf{96.29} & \textbf{86.11} & \textbf{81.68} & \textbf{142.32} & \textbf{5.99}\\ \bottomrule
\end{tabular}
}
\label{table1}
\vspace{-0.4cm}
\end{table*}

\subsection{Evaluation Metrics} 
To evaluate the model performance, we use the following key metrics: accuracy ($\alpha_{\text{acc}}$), mean accuracy ($\beta_{\text{macc}}$), frames per second (FPS), and total parameter count (Params). $\beta_{\text{macc}}$ represents the average accuracy across multiple tasks, providing an overall performance measure for multi-task learning. It is computed as:
\begin{equation}
\beta_{\text{macc}} = \frac{1}{m} \sum_{r=1}^{m} \alpha_{\text{acc}}^r,
\end{equation}
where and $\alpha_{\text{acc}}^r$ represents the accuracy of the model on the $r$-th task. 

\subsection{Experimental Details}
All experiments were conducted on an NVIDIA L40S GPU with 48GB VRAM. Training was performed for 125 epochs, with stabilization of the outcomes guiding early stopping. The initial learning rate was set to 0.001 and dynamically adjusted: reduced to 0.0005 between epochs 25 and 50, and further decreased to 0.00005 after epoch 50. The batch size was 24, with Stochastic Gradient Descent (SGD) used as the optimization algorithm, featuring a momentum of 0.9 and weight decay of 0.0001 to ensure convergence while mitigating overfitting.

\subsection{Comparison with SOTA Models}

We compared our model with state-of-the-art methods, following the experimental setup from \cite{yang2023aide}, which involves independent feature extraction from multimodal data and multi-view images. Our model consistently outperformed these baseline models across all evaluation metrics, including task-specific $\alpha_{\text{acc}}$, $\beta_{\text{macc}}$, FPS, and parameter count. With fewer than 6M parameters, our model improved $\beta_{\text{macc}}$ by 3.48\%-9.32\% and achieved an inference speed of 142.32 FPS, significantly surpassing the real-time requirements of ADAS systems.

These performance improvements are attributed to the model's design optimizations. Specifically, the joint feature extraction across similar modalities reduces model complexity and enhances inter-modal interactions, while the adaptive weighting strategy addresses task-specific feature importance, mitigating negative transfer and improving scalability. These innovations enable our model to achieve superior results in terms of both performance and efficiency, demonstrating its practical applicability for ADAS.

\subsection{Ablation Experiment}
We conducted a series of ablation experiments to evaluate the individual contributions of the MTS-Mamba and MGMI modules, as well as their key components, and to investigate the interactive effects between tasks and multimodal data.

\begin{table}[t]
\setlength{\tabcolsep}{7pt}
\centering
\caption{Ablation experiment results of MTS-Mamba and MGMI. "w/" indicates that the corresponding component is used, while "w/o" denotes that the component is not used.}
\resizebox{\linewidth}{!}{%
\begin{tabular}{cc|cccc|c|c|c}
\toprule                                              
 & & \multicolumn{4}{c|}{$\alpha_{\text{acc}} (\%) \uparrow $} & \multirow{2}{*}{\centering \begin{tabular}[c]{@{}c@{}}$\beta_{\text{macc}}$ \\ $(\%) \uparrow$\end{tabular}} & \multirow{2}{*}{\centering FPS $\uparrow$} & \multirow{2}{*}{\centering P(M) $\downarrow$} \\ \cline{3-6}
 \multirow{-2}{*}{\centering \begin{tabular}[c]{@{}c@{}}MTS- \\ Mamba\end{tabular}} & \multirow{-2}{*}{\centering MGMI} & \raisebox{-0.6ex}{\centering \textsc{der}}  & \raisebox{-0.6ex}{\centering \textsc{dbr}} & \raisebox{-0.6ex}{\centering \textsc{tcr}}  & \raisebox{-0.6ex}{\centering \textsc{vbr}} & &  &  \\ 
\midrule

 w/o  &  w/o  & 67.38 & 58.75 & 83.06 & 69.38 & 69.64  & 101.83 & 27.68  \\
 w/   &  w/o  & 72.13 & 64.57 & 92.05 & 77.16 & 76.48  & \textbf{158.84} & \textbf{5.73}  \\
 w/o  &  w/   & 70.54 & 62.31 & 90.37 & 73.84 & 74.02  & 89.43 & 28.12 \\ \midrule
\rowcolor{gray!15} 
w/   &  w/   & \textbf{75.00} & \textbf{69.31} & \textbf{96.29} & \textbf{86.11} & \textbf{81.68} & 142.32 & 5.99\\

\bottomrule
\end{tabular}
}
\label{table2}
\end{table}

\subsubsection{Ablation experiments on MTS-Mamba and MGMI}

We designed three experimental configurations to assess the contributions of the MTS-Mamba and MGMI modules. In the first group, we replaced MTS-Mamba with a simple VGG16 network and MGMI with basic concatenation fusion. The results, shown in Table \ref{table2}, demonstrate significant improvements in $\beta_{\text{macc}}$ by 5.2\%-12.04\% when both MTS-Mamba and MGMI were used. Specifically, replacing VGG16 with MTS-Mamba improved $\beta_{\text{macc}}$ by 7.66\% and increased FPS by 52.89, while reducing parameters by over four times. Similarly, replacing concatenation fusion with MGMI resulted in a 5.2\% increase in $\beta_{\text{macc}}$ with only an additional 0.26M parameters. These findings highlight the effectiveness of both MTS-Mamba and MGMI in efficiently extracting temporal-spatial features and mitigating task conflicts in multi-task learning (MTL).

We also evaluated the contribution of the forward-backward temporal scanning mechanisms and the global-local spatial feature extraction module in MTS-Mamba. When either of these components was removed, $\beta_{\text{macc}}$ dropped by 3.47\%-3.82\%, as shown in Table \ref{table3}. This decline underscores the importance of the synergy between these components in extracting rich spatiotemporal features and synchronizing bidirectional temporal information, providing robust representations for multi-task recognition. Additionally, despite these performance gains, the model maintained its inference speed and did not introduce additional parameters, confirming the efficiency of the MTS-Mamba design.

\subsubsection{Ablation on Self-Attention and Multi-Gating Mechanisms in MGMI}
We analyzed the role of the self-attention and multi-gating mechanisms in MGMI. Removing either of these components led to a noticeable decline in $\beta_{\text{macc}}$, especially when the multi-gating mechanism was excluded (Table \ref{table4}). This result is expected, as the multi-gating mechanism dynamically adjusts the fusion weights of each modality's features according to the task's requirements, enabling selective emphasis on the most relevant modality for each task. This adjustment alleviates feature conflicts in MTL and significantly improves the model's performance across multiple tasks. The experimental results reinforce the importance of both the self-attention and multi-gating mechanisms in enhancing feature fusion for multi-task learning.

\begin{table}[t]
\setlength{\tabcolsep}{7pt}
\centering
\caption{Ablation experiment results of the forward-backward temporal scanning mechanisms and the global-local spatial feature extraction module in MTS-Mamba.}
\vspace{-0.2cm}
\resizebox{\linewidth}{!}{%
\begin{tabular}{cc|cccc|c|c|c}
\toprule                                              
 & & \multicolumn{4}{c|}{$\alpha_{\text{acc}} (\%) \uparrow $} & \multirow{2}{*}{\centering \begin{tabular}[c]{@{}c@{}}$\beta_{\text{macc}}$ \\ $(\%) \uparrow$\end{tabular}} & \multirow{2}{*}{\centering FPS $\uparrow$} & \multirow{2}{*}{\centering P(M) $\downarrow$} \\ \cline{3-6}
 \multirow{-2}{*}{\centering \begin{tabular}[c]{@{}c@{}}Global- \\ Local\end{tabular}} & \multirow{-2}{*}{\centering \begin{tabular}[c]{@{}c@{}}Dual-path  \\ Scanning\end{tabular}} & \raisebox{-0.6ex}{\centering \textsc{der}}  & \raisebox{-0.6ex}{\centering \textsc{dbr}} & \raisebox{-0.6ex}{\centering \textsc{tcr}}  & \raisebox{-0.6ex}{\centering \textsc{vbr}} & &  &   \\ 
\midrule
 % w/o  &  w/o  &  62.91     & 60.73     & 82.84      & 74.33      & 70.25  \\
 w/   &  w/o  & 73.15      & 66.85     & 91.21      & 80.22     & 77.86  & 146.59 & \textbf{5.99}  \\
 w/o  &  w/   & 72.78     & 67.30      & 92.11      & 80.65   & 78.21  & \textbf{148.84} & \textbf{5.99} \\ \midrule
\rowcolor{gray!15} 
w/   &  w/   & \textbf{75.00} & \textbf{69.31} & \textbf{96.29} & \textbf{86.11} & \textbf{81.68} & 142.32 & \textbf{5.99}\\

\bottomrule
\end{tabular}
}
\label{table3}
\end{table}

\begin{table}[t]
\setlength{\tabcolsep}{7pt}
\centering
\caption{Ablation experiment results of the self-attention and multi-gating mechanisms in MGMI.}
\vspace{-0.2cm}
\resizebox{\linewidth}{!}{%
\begin{tabular}{cc|cccc|c}
\toprule                                              
 & & \multicolumn{4}{c|}{$\alpha_{\text{acc}} (\%) \uparrow $} & \multirow{2}{*}{\centering \begin{tabular}[c]{@{}c@{}}$\beta_{\text{macc}}$ \\ $(\%) \uparrow$\end{tabular}} \\ \cline{3-6}
 \multirow{-2}{*}{\centering \begin{tabular}[c]{@{}c@{}}Self- \\ Attention\end{tabular}} & \multirow{-2}{*}{\centering \begin{tabular}[c]{@{}c@{}}Multi-gating \\ Mechanism\end{tabular}} & \raisebox{-0.6ex}{\centering \textsc{der}}  & \raisebox{-0.6ex}{\centering \textsc{dbr}} & \raisebox{-0.6ex}{\centering \textsc{tcr}}  & \raisebox{-0.6ex}{\centering \textsc{vbr}} & \\ 
\midrule

 w/o & w/o & 72.13 & 64.57 & 92.05 & 77.16 & 76.48 \\
 w/ &  w/o & 74.60 & 65.84 & 93.29 & 84.75 & 79.62 \\
 w/o &  w/ & 73.53 & 64.92 & 91.99 & 82.39 & 78.21 \\ \midrule
\rowcolor{gray!15} 
w/   &  w/   & \textbf{75.00} & \textbf{69.31} & \textbf{96.29} & \textbf{86.11} & \textbf{81.68} \\

\bottomrule
\end{tabular}
}
\label{table4}
\end{table}

\subsubsection{Ablation experiments between different tasks}

To explore the advantages of MTL and the interactions between different tasks, we designed a series of ablation experiments. The four tasks were grouped into two dimensions: driver state recognition (\textsc{der} and \textsc{dbr}) and traffic environment recognition (\textsc{tcr} and \textsc{vrb}). We performed two sets of experiments (Table \ref{table5}): we retained only the driver state recognition tasks (\textsc{der} and \textsc{dbr}), excluding the traffic environment tasks (\textsc{tcr} and \textsc{vbr}). The results showed a drop in $\alpha_{\text{acc}}$ for \textsc{der} and \textsc{dbr} by 1.86\%-2.13\%. In the second set, we retained only the traffic environment recognition tasks (\textsc{tcr} and \textsc{vbr}), excluding the driver state classification tasks (\textsc{der} and \textsc{dbr}), which resulted in a more significant $\alpha_{\text{acc}}$ drop of 4.82\%-5.86\%. These results demonstrate that the tasks within the MTL framework benefit from significant synergies. Jointly learning both driver state and traffic environment recognition tasks enhances the model's overall accuracy and generalization capability.

\begin{table}[t]
\def\arraystretch{1.25}
\caption{Ablation experimental results for driver state recognition tasks (i.e., \textsc{der}, \textsc{dbr}) and traffic environment recognition tasks (i.e., \textsc{tcr}, \textsc{vbr}).}
\resizebox{\linewidth}{!}{%
\begin{tabular}{cccc|cccc}
\toprule
\multicolumn{4}{c|}{\raisebox{0.5ex}{\centering Task}} & \multicolumn{4}{c}{\raisebox{0.5ex}{\centering $\alpha_{\text{acc}} (\%) \uparrow $}} \\ \cline{1-8} 
\multicolumn{2}{c}{\raisebox{-0.5ex}{\centering Driver States}} & 
\multicolumn{2}{c|}{\raisebox{-0.5ex}{\centering Traffic Environment}} & 
\raisebox{-0.5ex}{\centering \textsc{der}}  & \raisebox{-0.5ex}{\centering \textsc{dbr}} & \raisebox{-0.5ex}{\centering \textsc{tcr}}  & \raisebox{-0.5ex}{\centering \textsc{vbr}} \\ 
\midrule

 \multicolumn{2}{c}{w/} &\multicolumn{2}{c|}{w/o} &  73.14 & 67.18 & - & - \\

 \multicolumn{2}{c}{w/o}& \multicolumn{2}{c|}{w/} & - & - & 91.47 & 80.25 \\ \midrule

\rowcolor{gray!15} \multicolumn{2}{c}{w/} & \multicolumn{2}{c|}{w/}  & \textbf{75.00} & \textbf{69.31} & \textbf{96.29} & \textbf{86.11} \\ \bottomrule

\end{tabular}
}
\label{table5}
\end{table}

\subsubsection{Ablation experiments on multimodal data}
\label{am}

We validate the independent contributions of each modality through ablation experiments by categorizing the modalities into three groups: vehicle-exterior images (front-view, left-view, right-view), vehicle-interior images (inside-view, driver’s facial and body images), and joint data (posture and gesture). We trained the model with each data group separately, and the results are shown in Table \ref{table6}.

The results demonstrate that models trained with a single modality perform worse than the multimodal model, with a drop of 5.82\% to 14.18\% in $\beta_{\text{macc}}$. This confirms the crucial role of multimodal data in ADAS-related tasks. Further analysis reveals that different modalities benefit different tasks: vehicle-exterior images improve accuracy for \textsc{tcr} and \textsc{vbr}, while vehicle-interior images and joint data enhance the  accuracy of \textsc{der} and \textsc{dbr} tasks. This variation arises because each modality expresses different information—vehicle-exterior images reflect road conditions, while vehicle-interior images and joint data capture the driver’s behavior and facial expressions. These findings validate the necessity of our MGMI design, which adaptively adjusts the weights of different modalities to alleviate negative transfer and improve task performance in MTL scenarios.

\begin{table}[t]
\setlength{\tabcolsep}{7pt}
\renewcommand{\arraystretch}{1.2}
\centering
\caption{Results of the Ablation Experiments on Multimodal Data.}
\resizebox{\linewidth}{!}{%
\begin{tabular}{ccl|cccc|c}
\toprule
& & & \multicolumn{4}{c|}{$\alpha_{\text{acc}} (\%) \uparrow $} & \multirow{2}{*}{\centering \begin{tabular}[c]{@{}c@{}}$\beta_{\text{macc}}$ \\ $(\%) \uparrow$\end{tabular}} \\ \cline{4-7}
 \multirow{-2}{*}{\centering \begin{tabular}[c]{@{}c@{}}Vehicle-exterior \\ Images\end{tabular}} & \multirow{-2}{*}{\centering \begin{tabular}[c]{@{}c@{}}Vehicle-interior \\  Images\end{tabular}} & \multirow{-2}{*}{\centering Joints} & \raisebox{-0.4ex}{\centering \textsc{der}}  & \raisebox{-0.4ex}{\centering \textsc{dbr}} & \raisebox{-0.4ex}{\centering \textsc{tcr}}  & \raisebox{-0.4ex}{\centering \textsc{vbr}} &\\ 
\midrule

\CheckmarkBold & & & 69.78 & 60.27 & 92.71 & 80.68 & 75.86 \\
& \CheckmarkBold & & 71.23 & 61.49 & 84.86 & 72.03 & 72.40 \\
& & \multicolumn{1}{c|}{\CheckmarkBold} & 70.39 & 65.53 & 73.33 & 60.74 & 67.50 \\ \midrule
\rowcolor{gray!15} 
 \CheckmarkBold & \CheckmarkBold & \multicolumn{1}{c|}{\CheckmarkBold} & \textbf{75.00} & \textbf{69.31} & \textbf{96.29} & \textbf{86.11} & \textbf{81.68} \\ 
 \bottomrule
\end{tabular}
}
\label{table6}
\vspace{-0.4cm}
\end{table}

\section{Conclusion}
This paper introduces a TEM$^{3}$-Learning (Time-Efficient Multimodal Multi-Task Learning) framework designed to recognize driver emotion, behavior, traffic context, and vehicle behavior simultaneously. TEM$^{3}$-Learning integrates two key components: MTS-Mamba, which efficiently captures temporal-spatial features from multi-view sequential images, and MGMI, which adaptively adjusts the weights of modality features for each task using a multi-gate mechanism. This design alleviates negative transfer between tasks, optimizing performance across multiple recognition tasks. Experimental results on the AIDE dataset demonstrate that TEM$^{3}$-Learning achieves superior performance in all four recognition tasks, with an inference speed exceeding the baseline models, while maintaining fewer than 6 million parameters. These findings highlight the efficiency, scalability, and practical applicability of TEM$^{3}$-Learning in real-time ADAS systems. TEM$^{3}$-Learning and its core components offer a valuable contribution to multimodal multi-task learning in ADAS, paving the way for the development of more efficient and robust algorithms in this field.

\bibliographystyle{IEEEtran}
\bibliography{TotalBib}

\end{document}